\pdfoutput=1

\documentclass[11pt]{article}

\usepackage[final]{acl}

\usepackage{times}
\usepackage{latexsym}
\usepackage{colortbl}
\usepackage{booktabs}
\usepackage{enumitem}
\usepackage{titlesec}
\usepackage{float}
\usepackage{multicol}
\usepackage{url}
\usepackage{hyperref}

\usepackage[T1]{fontenc}
\usepackage[colorinlistoftodos]{todonotes}

\usepackage[utf8]{inputenc}

\usepackage{microtype}

\usepackage{inconsolata}

\usepackage{graphicx}

\usepackage[export]{adjustbox}
\title{\textsc{Autalic}: A Dataset for Anti-AUTistic Ableist Language In Context}

\author{
  \textbf{Naba Rizvi\textsuperscript{1}},
  \textbf{Harper Strickland\textsuperscript{1}},
  \textbf{Daniel Gitelman\textsuperscript{1}},
  \textbf{Tristan Cooper\textsuperscript{1}},\\
  \textbf{Alexis Morales-Flores\textsuperscript{1}},
  \textbf{Michael Golden\textsuperscript{1}},
  \textbf{Aekta Kallepalli\textsuperscript{1}},
  \textbf{Akshat Alurkar\textsuperscript{1}},\\
  \textbf{Haaset Owens\textsuperscript{1}},
  \textbf{Saleha Ahmedi\textsuperscript{1}},
  \textbf{Isha Khirwadkar\textsuperscript{1}},
  \textbf{Imani Munyaka\textsuperscript{1}},\\
  \textbf{Nedjma Ousidhoum\textsuperscript{2}}
  \\ \\
\textsuperscript{1} University of California, San Diego,
 \textsuperscript{2} Cardiff University,
\\
 \small{
   \textbf{Correspondence:} \href{mailto:nrizvi@ucsd.edu}{nrizvi@ucsd.edu}
 }
}


\begin{document}
\maketitle
\begin{abstract}
As our awareness of autism and ableism continues to increase, so does our understanding of ableist language towards autistic people. Such language poses a significant challenge in NLP research due to its subtle and context-dependent nature. Yet, detecting anti-autistic ableist language remains underexplored, with existing NLP tools often failing to capture its nuanced expressions. 
We present \textbf{\textsc{Autalic}}, the first dataset dedicated to the detection of anti-autistic ableist language in context, addressing a significant gap in the field. \textbf{\textsc{Autalic}} comprises 2,400 autism-related sentences collected from Reddit, accompanied by surrounding context, and annotated by trained experts with backgrounds in neurodiversity. Our comprehensive evaluation reveals that current language models, including state-of-the-art LLMs, struggle to reliably identify anti-autistic ableism and diverge from human judgments, underscoring their limitations in this domain. We publicly release our dataset\footnote{\url{https://nrizvi.github.io/AUTALIC.html}} along with the individual annotations, providing an essential resource for developing more inclusive and context-aware NLP systems that better reflect diverse perspectives.
\end{abstract}

\textcolor{red}{Trigger warning: this paper contains ableist language including explicit slurs and references to violence.}
\section{Introduction}

\begin{figure}[ht]
    \centering
    \includegraphics[width=\columnwidth]{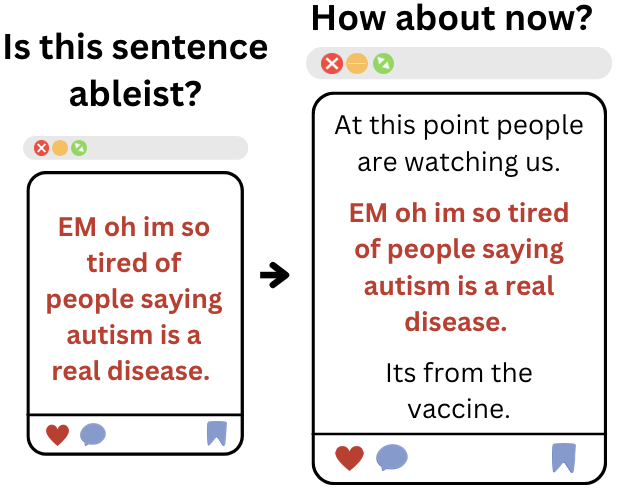}
    \captionsetup{hypcap=false} 
    \captionof{figure}{
      The example illustrates the importance of labeling sentences in context. The target sentence alone, shown on the left, is difficult to classify as ableist toward autistic people. Adding the surrounding sentences, as shown on the right, provides context revealing the original poster's reference to the debunked vaccine-autism stereotype, which is tied to anti-autistic stigma~\cite{mann2019autism,davidson2017vaccination}.
    }
    \label{fig:teaser}
\end{figure}

There are several critical frameworks used to define autism~\cite{lawson2021social}, including the medical model, which defines disability as a ``disease'' and remains one of the most widely used in computer science research focusing on autism~\cite{spiel_agency_2019, anagnostopoulou_artificial_2020, parsons_whose_2020, sideraki_artificial_2021,sum_dreaming_2022,williams_counterventions_2023, rizvi2024}. 
Since this framework defines autism as a deficit of skills, its applications in technology research largely focus on providing diagnosis and treatment to autistic people~\cite{baron1997mindblindness, begum2016robots,spiel2019,rizvi2024}. This belief also posits neurotypical behaviors as the ``norm'' and autism as a ``deficit'' of these norms, thereby promoting neuronormativity instead of neurodiversity, which views all neurotypes as valid forms of human diversity~\cite{bottema2021avoiding, walker2014neurodiversity}.

To move beyond the limitations of deficit-based models and better align NLP and AI research with neurodiversity, we present \textbf{\textsc{Autalic}}, a dataset of 2,400 autism-related sentences collected from Reddit, accompanied by 2,014 preceding and 2,400 following context sentences. The dataset addresses a critical gap in current NLP research, which has largely overlooked the nuanced and context-dependent nature of ableist speech targeting autistic individuals. \textbf{\textsc{Autalic}} captures key contextual elements and includes annotations by trained experts familiar with the autistic community, ensuring higher reliability and relevance. As \textbf{\textsc{Autalic}} contains all individual labels to capture the nuances in human perspectives, it can serve as a resource for researchers studying anti-autistic ableist speech, neurodiversity, or disagreements in general.

Through a series of experiments with different baselines including 4 LLMs, our evaluation indicates that while some LLMs such as DeepSeek have the most consistent scores regardless of the language used in the prompt, thereby indicating a more thorough understanding of the different ways anti-autistic speech may be identified or may manifest in text, they remain unreliable agents for annotating such a task. We find that in-context learning examples provide mixed results in helping improve the task comprehension among LLMs. Our results provide empirical evidence highlighting the difficulty of this task.

\section{Related Work}
Anti-autistic ableist language can be diverse in scope. It may include perpetuating stereotypes, using offensive language and slurs, or centering non-autistic people over the perspectives of autistic people~\cite{bottema2021avoiding, darazsdi2023oh, rizvi2024}. While abusive language detection systems can help identify such speech, they are known to demonstrate bias~\cite{manerba2021fine, venkit-etal-2022-study}, with even LLMs perpetuating ableist biases~\cite{gadiraju_offensive_23}. Additionally, anti-autistic ableist speech remains understudied, which is concerning given that classifiers trained on multiple hate speech datasets have shown a failure to generalize to target groups outside of the training corpus~\cite{yoder-etal-2022-hate}.

Although the language used to describe autism varies, prior studies with autistic American adults found 87\% prefer identity-first language over person-first language~\cite{taboas2023preferences}. Person-First Language (\textbf{PFL}) centers the person (e.g. ``person with autism''), while Identity-First Language (\textbf{IFL}) centers the identity (e.g. ``autistic person'')~\cite{taboas2023preferences}. Supporting this finding, other researchers have found that viewing autism as an identity may increase the psychological well-being of autistic individuals and lower their social anxiety~\cite{cooper2023impact}.

Ableist language online varies, may manifest in different ways and is ever-evolving ~\cite{welch2023understanding,heung2024vulnerable}. However, toxic language datasets focusing on hate speech and abusive language have often addressed disability in general terms but have not explicitly focused on autism~\cite{elsherief2018hate,ousidhoum2019multilingual}. To our knowledge, there are no previous datasets specifically focused on anti-autistic speech classification, and only 3 of the 23 datasets for bias evaluation in LLMs focus on disability~\cite{gallegos2024bias}. LLMs may be limited in that they lack an acknowledgment of context, which leads to higher rates of false positives when classifying ableist speech~\cite{phutane2024toxicity}. These limitations are also found in toxicity classifiers, which mainly identify explicit ableist speech but may otherwise perpetuate harmful social biases leading to content suppression~\cite{phutane2024toxicity}. 
Toxic language detection models, including LLM-based models, have been found to exhibit strong negative biases toward disabilities by classifying any disability-related text as toxic~\cite{narayanan-venkit-etal-2023-automated}. Further, LLMs have been observed to perpetuate implicitly ableist stereotypes \cite{disabilityllms}
and bias~\cite{gamaartificially, venkit-etal-2022-study}.
This, unfortunately, can sometimes be due to a research design that overlooks intra-community and disabled people's perspectives \cite{mondal-etal-2022-disabledonindiantwitter}, as well as autistic people's views, which may lead to harmful stereotypes~\cite{spiel2019,rizvi2024}. We make a step towards addressing these issues by building a dataset that focuses on ableist speech and autism and including autistic people's perspectives during the annotation process as recommended by~\cite{davani-etal-2023-hate}. \textbf{\textsc{Autalic}} contains all its labels and will also be useful for researchers interested in leveraging disagreements for difficult classification tasks~\cite{pavlick2019inherent,leonardelli2021agreeing}.

\section{\textsc{Autalic}}

To build \textbf{\textsc{Autalic}}, we collected relevant sentences containing autism-related keywords from Reddit using the methods described in Section \ref{subsec:collection}. The collected sentences were labeled by trained annotators, as discussed in Section \ref{subsec:annotation}.

\subsection{Data Collection}\label{subsec:collection}
We identify a methodology for curating sentences related to autism similar to prior datasets by collecting English-language sentences from Reddit ~\cite{d1, d2, d3, d4, d5, d6, d7}. 

\subsubsection{Data Collection Criteria}
We selected Reddit as our data source due to its popularity, emphasis on text-based content, and fewer API restrictions compared to X (formerly Twitter) at the time of data collection in January 2024. We performed keyword searches using Reddit’s default search settings, which prioritize relevance based on factors such as the rarity of terms in the query, the post's age, and its number of upvotes and comment.~\footnote{https://support.reddithelp.com/hc/en-us/articles/19695706914196-What-filters-and-sorts-are-available} The search terms include “autis*”, “ASD”, “aspergers”, and “disabilit*”; the full list is available in the Appendix, Table~\ref{tab:searchterms}.


\begin{figure}
    \centering
    \includegraphics[width=0.65\linewidth]{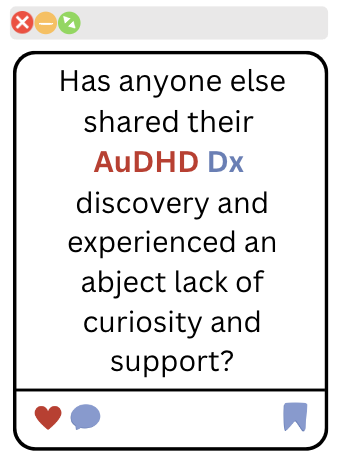}
\caption{An example sentence from our dataset. The search keyword is shown in \textcolor{red}{red}, while the word in \textcolor{blue}{blue} is an example of a term defined in our glossary.}
    \label{fig:words}
\end{figure}

We use the identified search terms to collect target sentence instances containing our keywords, which are then labeled by our annotators. To provide additional context, we also collect the sentences immediately preceding and following each target instance.
In total, we collect 2,400 target sentences, along with 2,014 preceding and 2,400 following sentences. We then split the dataset into three parts by randomly selecting and assigning 800 unique target sentences to each of three annotation segments, with each segment labeled by a group of three annotators.

\begin{table}
    \centering
    \begin{tabular}{cc} \hline
       \bfseries SubReddit  & \bfseries Sentence Count \\ \hline
         r/Aspergers & 116 \\ 
         r/Autism & 88 \\  
         r/AmITheAsshole & 39 \\ 
         r/AutisminWomen & 37 \\ 
         r/AuDHDWomen & 24 \\ \hline
    \end{tabular}
    \caption{The subreddits with the most sentences included in the \textbf{\textsc{Autalic}} dataset and the number of sentences extracted from each.}
    \label{tab:subreddits}
\end{table}
The average number of likes on each post included in the \textbf{\textsc{Autalic}} dataset is 1,611.59. Table~\ref{tab:subreddits} details the subreddits from which the most significant number of sentences were extracted from. With the exception of r/AmITheAsshole, all of the other subreddits are autism-related.

\subsubsection{Data Curation}\label{subsubsec:curation}

As some of the identified keywords may appear in other contexts, we perform an exact word search for the acronyms to ensure unrelated words that might contain our acronyms are excluded from the search as they go beyond the scope of our dataset. For example, we searched for ``applied behavioral analysis'' and a case-sensitive search for ``ABA'', which is a form of therapy intended to minimize autistic behaviors such as stimming (which is often used for self-soothing)~\cite{sandoval2019much}. Similarly, we exclude any posts that are not written in English using the Python package \texttt{langdetect} and posts that contain images, videos, or links.~\footnote{https://pypi.org/project/langdetect/}

\subsubsection{Final Dataset}\label{subsubsec:finalDataset}

Our final dataset includes 2,400 sentences from 192 different subreddits. To protect our annotators' privacy, we have anonymized individual label selections.

While nearly a quarter of the posts in our dataset were published in 2023, the range of publication years is 2013-2024. Figure~\ref{fig:words} shows an example of a sentence from our dataset that uses both a search keyword and a word defined in our glossary described in Section \ref{subsubsec:training}.

\subsection{Data Annotation}\label{subsec:annotation}

\subsubsection{Annotator Selection}\label{subsubsec:selection}
We recruit nine upper level undergraduate researchers as volunteer annotators and randomly assign them to annotate different segments of the dataset. We ensure that their involvement in our annotation process is voluntary, informed, and mutually beneficial. In particular, some participants chose to volunteer because they care deeply about the subject matter and wish to contribute to improving AI for autistic people. We select participants for whom this collaboration would provide relevant and valuable professional experience, grant them opportunities to engage in other aspects of the research, and provide mentorship and authorship recognition in accordance with the ACL guidelines. 

We prioritize the well-being and autonomy of our annotators by providing full disclosure of the research process and subject material prior to their participation. We supply relevant trigger warnings, discuss the nature of the content in detail during an orientation session, and allow annotators to make an informed decision about whether they wish to proceed. We make them aware that they are free to withdraw from the study at any point. 

Our annotators are US-based, culturally diverse, and include people who grew up outside the US. They are all fluent in English. Four of our annotators are gender minorities, and at least three self-identify as neurodivergent. Although we ensure the annotators were from diverse backgrounds during our recruitment process, due to the collaborative nature of our annotation process, we do not share the individual details of their identities. We also note that any personally identifiable information was destroyed upon the conclusion of our analysis and not shared outside of our research team. 

\subsubsection{Annotator Training}\label{subsubsec:training}

We provide a virtual orientation to all annotators explaining the history of anti-autistic ableism, examples of contemporary anti-autistic discrimination, and a brief overview of the annotation task.

The orientation begins with a discussion of the medical model approach to autism and its link to the Nazi eugenics program~\cite{waltz2008autism, sheffer2018asperger}. We define \textbf{neuronormativity} as the belief that the neurotypical brain is ``normal'' and other neurotypes are deficient in neurotypicality \cite{wise2023we}. We dive deeper into the medical model by discussing its impact on the self-perceptions and inclusion of autistic people in our society, such as an increase in suicidal ideation and social isolation among autistic people who mask or hide their autistic traits ~\cite{cassidy2014suicidal, cassidy2018risk}. Then, we cover the shifts in perspectives that emerged due to disability rights activism ~\cite{rowland2015angry,cutler2019listening}, and define \textbf{neurodiversity} as the belief that all neurotypes are valid forms of human diversity ~\cite{walker2014neurodiversity}. 

To explain the annotation task, we provide examples of sentences similar to what they may encounter while annotating. For example, we discuss how the inclusion of ``at least'' alters the connotations of the following sentence:
\begin{quote}
    \textbf{At least} I am not autistic.
\end{quote}
With just a minor change, the sentence can have an ableist connotation as it implies relief in knowing one is not autistic, as if it is shameful or wrong. 

We also introduce our glossary to the annotators as a dynamic resource that can be altered as needed. This glossary contains words that may appear in autism discourse online that may not be commonly known to others. These include medical acronyms, slang, and references to organizations and resources commonly affiliated with the autistic community (such as Autism Speaks). An excerpt of our glossary is available in our Appendix. We conclude our orientation by providing a brief tutorial video demonstrating how to run the script that will guide each annotator through the annotation task.

\subsubsection{Data Labeling}\label{subsubsec:labeling}
After completing the training, we assign each of the three segments of the dataset to three randomly selected annotators. Each annotator is assigned 800 unique sentences, with a goal of completing 200 annotations each week over four weeks. Annotators select from three possible labels for each sentence: ``Ableist'', ``Not Ableist'', or ``Needs More Context''.

\textbf{Ableist (1): } We ask our annotators to select this label if a sentence contains ableist sentiments as defined by the Center for Disability Rights: \textit{``Ableism is a set of beliefs or practices that devalue and discriminate against people with physical, intellectual, or psychiatric disabilities and often rests on the assumption that disabled people need to be `fixed’ in one form or the other.''}\footnote{https://cdrnys.org/blog/uncategorized/ableism/}

\textbf{Not Ableist (0): } Annotators select this label for sentences that describe positive or neutral behaviors and attitudes regarding autism, or posts written by an autistic person reaching out for help and support. This includes individuals using medical terminology in a personal context (e.g., \textit{``I need therapy''}), intra-community discussions, and general discussions of medical processes (unrelated to neurodivergence). Some examples of statements labeled as not ableist are: \textit{``I am autistic''}, \textit{``As an autistic person, I think...''}

\textbf{Needs More Context (-1): }This label is used for sentences an annotator is unable to definitively categorize as ableist or not ableist even with the contextual sentences provided. This category includes text that is entirely unrelated to disabilities or remains ambiguous without additional context.
\paragraph*{}
The number of times our annotators assign each label is detailed in Table~\ref{tab:labels}. To calculate our agreement scores, we consolidated labels -1 and 0 together based on feedback from our annotators that unrelated sentences needing more context could be classified as not anti-autistic in a purely granular classification. 
While we use the majority label as the ground truth in our analysis, we release the individual labels from each annotator due to a growing interest in embracing disagreements for such classification tasks in NLP~\cite{plank2014linguistically,pavlick2019inherent,leonardelli2021agreeing,kralj2022handling, plank-2022-problem}. 

\begin{table}
    \centering
    \begin{tabular}{p{0.8cm}p{4cm}p{0.9cm}} \hline
         \bfseries Label & \bfseries Definition & \bfseries Count \\ \hline
         -1 & unrelated to autism or needs more context & 595 \\ 
         0 & not ableist & 5,582 \\ 
         1 & ableist & 1,023 \\ 
         \hline
    \end{tabular}
    \caption{An explanation of the labels used in our classification task and the resulting counts of each label from all 9 annotators combined.}
    \label{tab:labels}
\end{table}

Our dataset contains 2,400 sentences labeled as containing anti-autistic ableist language or not. The labels are obtained by calculating the mode from the three annotators of each data segment. Using this methodology, 242 target sentences contain examples of anti-autistic ableist language (10\% of total), and 2,160 sentences do not (90\% of total).

\subsubsection{Providing Context}\label{subsubsec:context}
While we provide additional sentences for context, the annotators are instructed to annotate the target sentence exclusively and only refer to the other sentences for additional context, such as determining whether the sentence is part of an intra-community discussion or the use of figurative speech (i.e., sarcasm). Figure~\ref{fig:teaser} provides an example of a target sentence in context.
That is, in this example, it is difficult to determine whether or not the writer had ableist intent, as it can be interpreted in multiple ways---they can be critiquing the medical model, as many autistic activists do, thereby making it non-ableist; or they could be genuinely promoting ableist misrepresentations. 

The contextual sentences help the annotators better understand the writer's intent. With these sentences, it is apparent that the writer is referring to the harmful and widely discredited association of vaccines with autism, which not only promotes anti-autistic ableism in society but also puts people's lives at risk by spreading disinformation about the benefits and harms of life-saving vaccines~\cite{gabis2022myth, taylor2014vaccines, hotez2021vaccines}. 

Throughout the annotation process, annotators can edit previous annotations based on new knowledge to account for changes in language usage and connotations and the annotators' dynamic understanding of ableism.

\subsubsection{Disagreements}\label{subsubsec:disagreements}
The average Fleiss's Kappa scores are 0.25. This score underlines the difficulty of our classification task, which is apparent from the findings of prior works~\cite{ousidhoum2019multilingual}, including a quantitative assessment of tag confusions that found the majority of disagreements are due to linguistically debatable cases rather than errors in annotation ~\cite{plank2014linguistically}. Examples of such cases are provided in our Appendix.

We analyze the sentences with the highest levels of disagreement in our dataset. In 100 of these posts, we observe:
\begin{enumerate}[noitemsep,nolistsep]
    \item a tendency to use the medical model terminology or stereotypes (in 48 cases),
    \item a need for additional context beyond the sentences we provided.
\end{enumerate}

\begin{figure}
    \centering
    \includegraphics[width=0.99\linewidth]{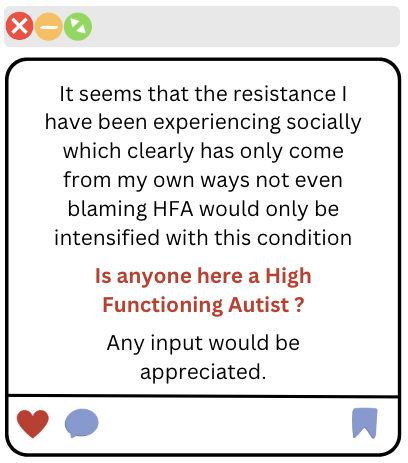}
    \caption{An example of a sentence from our dataset, shown in \textcolor{red}{red}, with a high level of disagreement among annotators.}
    \label{fig:disagreement}
\end{figure}
Figure~\ref{fig:disagreement} contains an example of a sentence with a high disagreement among our annotators. While functioning labels are considered ableist due to their eugenicist approach of categorizing autistic people based on their perceived economic value ~\cite{de2019binary}, it is difficult to determine whether the original poster is autistic or not. The context is important here as classifying a sentence such as this as ``ableist'' can lead to unfair censorship if the original poster is a self-diagnosed autistic person seeking advice. Therefore, these sentences were ultimately classified as ``not ableist'' in \textbf{\textsc{Autalic}}.

Our analysis reveals a moderately strong negative correlation between the task completion time and agreement with other annotators that is statistically significant ($R = -0.644$, $p$-value: $0.0096$). This highlights the importance of our orientation as we provided it simultaneously to the annotators. The annotators who completed their task immediately after our orientation had higher agreement.


\section{Experimental Setup}

We evaluate the performance of various baselines on our dataset using an 80\%–20\% train–test split. Additionally, we test four LLMs selected for their diverse range of model complexities.

\subsection{Baselines}
\paragraph{Logistic Regression (LR)}
We use logistic regression (LR) with Bag-of-Words (BoW) features for classification.

\paragraph{Fine-tuned BERT}
We fine-tune BERT~\cite{devlin2018bert} on our dataset to evaluate its performance on the classification task.

\subsection{Prompting LLMs}
We use four LLMs: Gemma2 ~\cite{team2024gemma}, Mistral ~\cite{jiang2023mistral}, Llama3 ~\cite{dubey2024llama}, and DeepSeek ~\cite{guo2025deepseek} to classify the sentences in our dataset. We adjust the prompts to compare each LLM's performance. Due to limitations in computational resources, we are unable to fine-tune the LLMs.

\paragraph{Prompts}
Due to the ever-evolving nature of language and variations in preferences among autistic individuals ~\cite{taboas2023preferences}, we use three different kinds of prompts to measure the consistency LLMs have in their understanding of anti-autistic ableist speech. These prompts include person-first (i.e., ``people with autism''), identity-first (i.e., ``autistic people''), and conceptual (i.e., ``anti-autistic'') language. 

We keep the default parameters for each LLM to maintain consistency, and prompt them with the following questions:
\begin{quote}For each target sentence, respond to the following questions with 0 for no or 1 for yes. Refer to the preceding and following sentences if more context is needed.
    \begin{itemize}
        \item Is this sentence ableist toward people with autism?
        \item Is this sentence anti-autistic?
        \item Is this sentence ableist toward autistic people? 
    \end{itemize}
\end{quote}

We include each sentence from \textbf{\textsc{Autalic}} after the aforementioned questions in our full prompt. In addition, we provide preceding and following context for each target sentence to the LLM to mimic the level of the information supplied to human annotators. 

\paragraph{Simple vs. ICL Prompting}
We run two sets of experiments with each LLM: one that uses zero-shot prompting, and another containing engineered prompts for in-context learning verbatim from the definitions and examples provided in our annotator orientation (Appendix Section~\ref{AO}).

\begin{table}[]
    \centering
    \begin{tabular}{p{2.0cm}p{1.2cm}p{1.2cm}p{0.9cm}}
    \toprule
    \multicolumn{4}{p{6cm}}{\centering \textit{Baselines}} \\ \hline
    \textbf{Model} & \textbf{Result} & \textbf{PT} & \textbf{FT}\\ \hline
    LR & 0.20 & -- & -- \\
    BERT & -- & 0.43 & 0.90 \\
    \midrule
    \multicolumn{4}{p{6cm}}{\centering \textit{Simple Prompting}} \\ \hline
    \textbf{LLM} & \textbf{PFL} & \textbf{IFL} & \textbf{AA} \\
    \hline
    Gemma2 & 0.23 & 0.19 & 0.33 \\
    Mistral & 0.28 & 0.27 & \textbf{0.34} \\
    Llama3 & 0.09 & 0.10 & \textbf{0.15} \\
    DeepSeek & 0.58 & 0.57 & \textbf{0.59} \\
    \hline
    \multicolumn{4}{p{6cm}}{\centering \textit{In-Context Learning}} \\ \hline
    Gemma2 & 0.25 & 0.24 & \textbf{0.34} \\
    Mistral & 0.31 & 0.24 & \textbf{0.34} \\
    Llama3 & 0.14 & 0.14 & 0.11 \\
    DeepSeek & 0.55 & 0.56 & 0.55 \\ \bottomrule
    \end{tabular}
    \caption{The F1 scores of various models using person-first language (PFL), identity-first language (IFL), and conceptual anti-autistic (AA) prompts with and without in-context learning examples for each LLM. The best scores for each model are in \textbf{bold}.}
    \label{tab:PromptE}
\end{table}

\subsection{Experimental Results}

\subsubsection{Fine-Tuning BERT}

The F1 scores for LR and BERT (both pretrained and fine-tuned) are presented in Table~\ref{tab:PromptE}, alongside results from all LLMs. Our experiments reveal that using BERT for this classification task can initially result in high rates of censorship if deployed as-is. The pretrained BERT model performed poorly, indicating its ineffectiveness in identifying anti-autistic ableist speech. In contrast, fine-tuning BERT on \textbf{\textsc{Autalic}} led to significant performance improvements across all metrics, enabling more accurate predictions of ableist speech, fewer false positives, and greater sensitivity to subtle expressions of ableism.

\subsubsection{Humans vs. LLMs}
\begin{figure}
    \centering
    \includegraphics[width=0.98\linewidth]{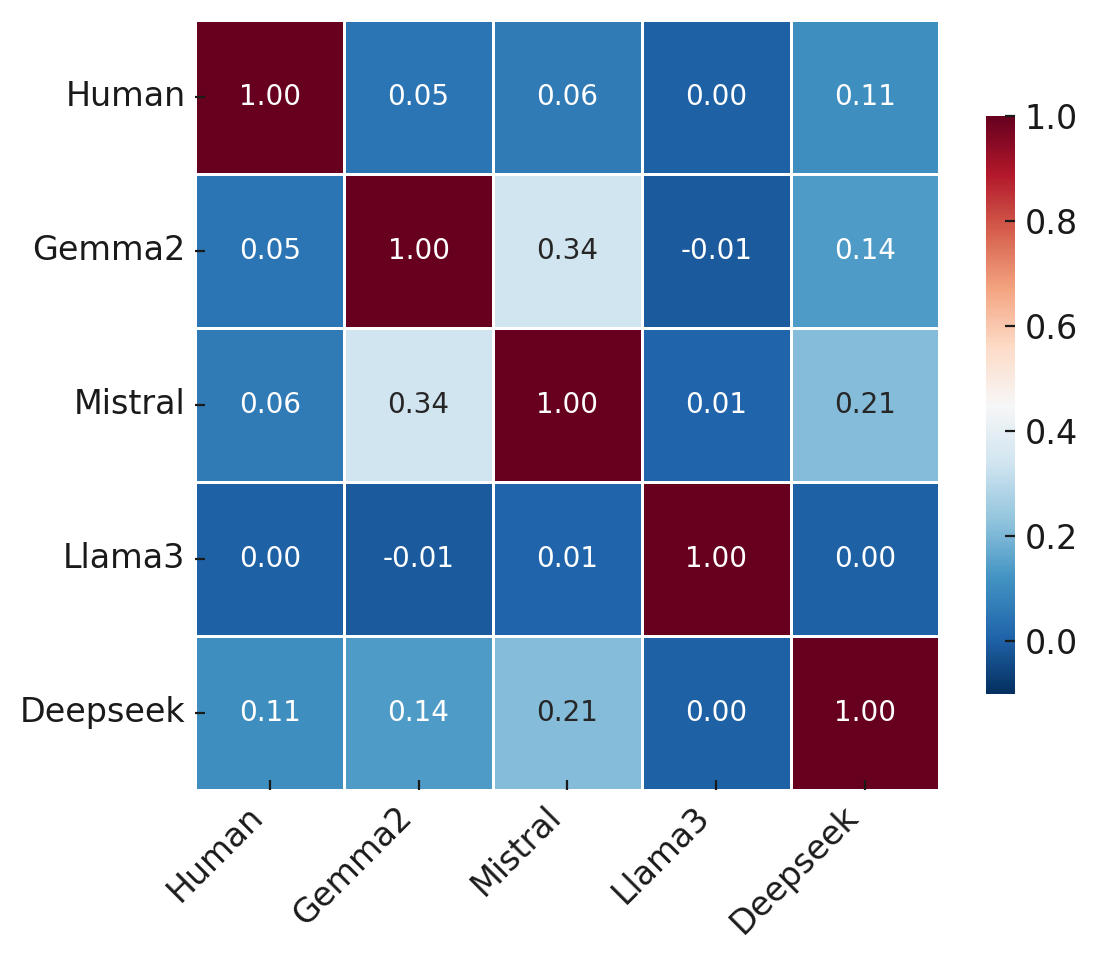}
    \caption{Mean Cohen’s Kappa scores for each LLM, comparing agreement with human annotators and other LLMs.}
    \label{fig:LLM_results}
\end{figure}

Our assessment reveals that LLMs show low levels of alignment with both human perspectives and the outputs of other LLMs, making them unreliable agents for this classification task if deployed as-is. We measure this alignment using Cohen's Kappa scores. As shown in Figure~\ref{fig:LLM_results}, the highest level of agreement was observed between Gemma2 and Mistral ($\kappa = 0.34$). No LLM demonstrated meaningful alignment with the human-annotated dataset, although DeepSeek’s agreement was notably higher than that of the other models. Overall, the LLMs exhibited low agreement with human judgments ($M = 0.091$, $SD = 0.110$). These results suggest that LLMs with fewer than 10 billion parameters struggle to classify anti-autistic ableist language, even when provided with in-context examples.

\subsubsection{In-Context Learning}
After providing in-context learning examples, Llama3 (+22.96\%) and Gemma2 (+12.68\%) show the largest relative improvements in F1 scores, indicating that both models benefit from the examples. In particular, our ICL examples also led to substantially more consistent scores across different prompt formulations for each LLM. For instance, supplying Llama3 with examples reduced its relative change in F1 score from 67.49\% to 17.40\% when switching from PFL to conceptual (AA) prompts, suggesting a better consideration of the connection between anti-autistic ableism and ableism toward autistic people.



\subsubsection{Identifying Ableist and Anti-Autistic Speech}
Our results with prompt engineering reveal that anti-autistic ableism is too abstract a concept for LLMs to reliably recognize, providing empirical evidence that they do not adequately reflect the perspectives of autistic people. LLMs struggle to identify anti-autistic speech regardless of the terminology used, further indicating that they are unreliable agents for this data annotation task.

Table~\ref{tab:PromptE} presents the results of prompt engineering using either person-first language (PFL), identity-first language (IFL), or the term ``anti-autistic'' (AA) to frame this form of ableism in more conceptual terms. Notably, switching from PFL or IFL to AA resulted in the largest relative changes in performance. For instance, prompting Llama with \textit{``Is this sentence \textbf{anti-autistic}?''} instead of \textit{``Is this sentence \textbf{ableist toward people with autism}?''} led to a relative increase of 67.49\% in F1-score. These results indicate that LLMs struggle to frame anti-autistic ableism and ableism toward autistic people as referring to the same phenomenon. Even after introducing ICL examples, the relative change in F1-scores between different prompt types remained high---for example, 32.66\% for Gemma2.





%
Interestingly, DeepSeek has the best results and highest consistency out of all the others. Although its agreement with human annotators is low ($k:0.11$), it is still double that of all other LLMs, as shown in Figure ~\ref{fig:LLM_results}. This highlights the difficulty of this task, as more advanced reasoning is required to understand the nuances of anti-autistic ableism, including the terminology we use to describe such speech.

\subsection{Error Analysis: Over-classification and Contextual Challenges for LLMs}
We analyze the top 10\% of sentences in which our human annotators were in perfect agreement, but the LLMs disagreed with them. In doing so, we find many instances where the LLMs over-classified sentences as ableist. For example, Llama labeled 42 sentences as ableist, compared to 0 by our human annotators. Of these 42 sentences, 74-93\% were also labeled as ableist by the other LLMs. Notably, 29 of these sentences involved intra-community discussions, highlighting the extent to which community voices may be censored if LLMs are used for content moderation.

\begin{figure}
    \centering
    \includegraphics[width=0.65\linewidth]{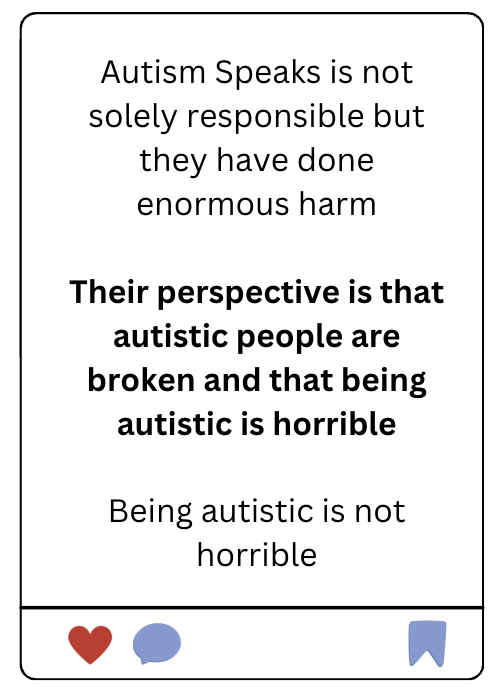}
    \caption{An example of a sentence that LLMs classified as ``anti-autistic'', in direct disagreement with all of our human annotators.}
    \label{fig:error_analysis}
\end{figure}

While two of the sentences did contain slurs, and 34 included other negatively connoted words---such as ``burden'', ``threat'', ``harm'', or ``stupid''---these terms were not used in an anti-autistic context. Figure \ref{fig:error_analysis} presents an example of such a sentence. Despite including negatively connoted terms, the sentence expresses an organization's viewpoint that the author explicitly disagrees with. Since our annotators were instructed to consider context, they correctly labeled this sentence as not ableist. However, the LLMs misclassified it.

This example demonstrates that even when given contextual information, LLMs tend to misclassify sentences based on the mere presence of negative words or connotations. This tendency contributes to their high rates of community censorship if deployed for moderation.

\subsection{Discussion}
Classifying anti-autistic ableist speech is challenging even within a Western, English-speaking context, as perceptions of what constitutes anti-autistic content differ significantly—even among autistic individuals \cite{keating2023autism}. Factors such as personal experience, cultural norms, and the evolving discourse around autism advocacy influence each individual’s perception of, and sensitivity to, toxic speech. This variability makes it difficult to establish a consistent classification scheme \cite{kapp2013deficit, ousidhoum2019multilingual, bottema-beutel_avoiding_2021, taboas2023preferences}.

While developing \textbf{\textsc{Autalic}}, we standardized our definition of anti-autistic ableist speech by providing annotators with an orientation and a glossary. These resources were created in alignment with the perspectives of autistic individuals \cite{bottema2021avoiding, taboas2023preferences, kapp2013deficit}.

Standard baselines such as logistic regression (LR) and BERT performed poorly, underscoring the need for domain-specific fine-tuning. Further, even after fine-tuning BERT on \textbf{\textsc{Autalic}}, our experiments reveal a high false positive rate, which could result in unfair censorship if used for content moderation. Furthermore, our results show that even state-of-the-art LLMs exhibit low agreement with human annotators, highlighting the difficulty of detecting subtle forms of ableism with general-purpose models. Among the models evaluated, DeepSeek achieved the best performance, further emphasizing the complexity of the task.

These findings demonstrate the importance of \textbf{\textsc{Autalic}} in aligning LLM performance with human expectations in the context of autism inclusion and ableist speech classification. Our experiments provide empirical evidence of the current limitations of using LLMs and traditional classifiers to identify anti-autistic ableism. These limitations include misalignment with human perspectives, a lack of conceptual understanding of anti-autistic ableism, and inconsistency---even when provided with in-context learning examples. Each of these challenges hinders the reliability of LLMs for this task.


\section{Conclusion}
In this paper, we introduced \textsc{Autalic}, the first dataset specifically focused on anti-autistic ableist language in context. By collecting and annotating 2,400 Reddit sentences along with their surrounding context, we aim to fill a critical gap in current NLP research and improve its alignment with the principles of neurodiversity.
\newline
By addressing bias against autistic individuals, \textsc{Autalic} marks a significant step forward in research on ableism within content moderation systems, hate speech detection models, and broader studies of neurodiversity and discrimination. We hope to support the development of more equitable NLP systems that better serve underrepresented and marginalized groups by sharing this resource with the research community,

\section{Limitations}
\label{limitations}
Despite our efforts to include a variety of ableist language targeting autistic individuals, our dataset may still exhibit some selection bias \cite{ousidhoum2020comparative}, as we relied on keyword searches and specific social media threads for data collection.

We acknowledge that, like many datasets presented at ACL, our work primarily reflects Western perspectives \cite{anderson-chavarria_autism_2022,CASE2022-MultimodalHate,aoyama2023gentle,SemEval2023-EDOS,10.1145/3555776.3577736,takeshita2024aclsum}. Therefore, we do not claim that \textbf{\textsc{Autalic}} is generalizable across languages and cultures, as anti-autistic and ableist speech may manifest and be perceived differently in different contexts. However, we emphasize that \textbf{\textsc{Autalic}} is the first dataset to center the perspectives of autistic people in ableist speech detection, helping to move away from the tendency of AI systems to misclassify any disability-related text as ``toxic'' \cite{hutchinson-etal-2020-social,narayanan-venkit-etal-2023-automated,VanDorpe2023Unveiling, heung2024vulnerable}. As such, it can serve as a valuable reference for future work exploring ableist speech in diverse linguistic and cultural contexts.

Our dataset consists of sentences sourced from Reddit, with many posts coming from autism-related subreddits such as r/Aspergers, r/Autism, and r/AutismInWomen. Our search criteria included terms like ``r*tard'', a slur broadly used against people with intellectual disabilities, not exclusively targeting autism. While we made efforts to filter out irrelevant posts, our inclusion of broader search terms like ``neurodiver'' (e.g., ``neurodiversity'', ``neurodiverse'', and ``neurodivergent'') also returned unrelated content, such as discussions about the NEURODIVER video game. Most of the dataset's content was published in 2023. Although our dataset is relatively small, the quality of the annotations is high, due to standardized protocols and resources developed through rigorous, iterative testing.
\section{Ethics Statement}
The data collected for our small-scale, non-commercial research complies with Reddit's API limits and policies \cite{RedditDataAPITerms2023, RedditUserAgreement2023}. All sentences in our dataset are publicly available, and our data collection process follows the methodologies established by prior work \cite{atuhurra2024revealing}.

We obtained IRB approval from our university's ethics review board and recruited volunteer annotators through our connections with various academic groups. Given the sensitivity of the content, we provided annotators with appropriate trigger warnings and allowed them to work at their own pace or withdraw from the study at any time. Additionally, we facilitated communication within each annotation team to support discussion of the content and the annotation process as needed. We specifically recruited annotators for whom this collaboration---and potential paper authorship---would be mutually beneficial, and we provided a comprehensive overview of the task to ensure informed consent. Some annotators chose to volunteer due to their strong commitment to neurodiversity and autism inclusion.

While our dataset and associated citations will be made available to the academic community, commercial use---including use by state actors in high-risk domains (e.g., healthcare)---is strictly prohibited. Furthermore, as our understanding of anti-autistic ableism continues to evolve, the classifications in \textbf{\textsc{Autalic}} may become outdated. We will include disclaimers to reflect such developments as needed, but we strongly encourage researchers building on our work to stay informed by consulting current perspectives from autistic scholars, activists, and organizations. Please refer to the Appendix for our Guidelines for Responsible Use.

\bibliography{custom}
\appendix
\section{Appendix}
\subsection{Guidelines for Responsible Use}
\subsubsection{Purpose and Scope}
This dataset has been curated to aid in the classification and study of anti-autistic ableist language in a U.S. context using text from Reddit. It aims to support research and educational endeavors focused on understanding, identifying, and mitigating ableist speech directed at autistic individuals, while moving away from mis-classifying any speech related to autism or disability as toxic.

\subsubsection{Applicability and Cultural Context}

\textbf{U.S.-Specific Results:} The language examples and classification models in this dataset are primarily reflective of usage and cultural nuances in the United States. As a result, the dataset and any models developed from it may not be fully accurate or generalizable for other countries or cultural contexts.

\textbf{Data Curation:} The data included in this has been taken solely from posts and comments on Reddit and may not represent autism discourse on other platforms and in other contexts.

\textbf{Contact for Latest Version:} Language evolves over time. For the most up-to-date version of the dataset, or on more information on when the dataset was last updated, please contact the first author.

\subsubsection{Access and Security}

\textbf{Password Protection:} The dataset is password-protected to prevent unauthorized or automated scraping (e.g., by bots). While the password is publicly available as of this publication, it may require prior approval in the future as needed to ensure reponsible use.

\textbf{Secure Storage:} Users are expected to maintain secure protocols (e.g., encryption, controlled access) to prevent unauthorized sharing or leaks of the dataset. The dataset may not be shared without consent of the authors. 

\subsubsection{Permitted Uses}

\textbf{Free Use for Scientific Research:} The dataset is publicly available without charge for legitimate scientific, academic, or educational research purposes, subject to the restrictions outlined below.

\textbf{Academic and Non-Profit Contexts:} Users in academic, research, or non-profit institutions may incorporate the dataset into studies, presentations, or scholarly articles, provided they follow these guidelines and appropriately cite the dataset and its authors.

\subsubsection{Prohibited or Restricted Uses
Commercial Use:} Commercial use is not authorized without explicit written permission from the dataset authors. If you wish to incorporate the dataset into commercial products or services, you must obtain approval in advance.

\textbf{Automated Content Moderation:} Using the dataset to develop or deploy automated content moderation tools is not authorized without prior approval from the authors. This restriction helps ensure that any moderation system is deployed ethically and with proper considerations for context and language evolution.

\subsubsection{Ethical Considerations and Privacy}

\textbf{Respect for Individuals and Communities:} Users must handle the dataset with an understanding of the impacts of ableist language on autistic communities. The dataset’s examples are provided solely for research and analysis and  must not be used to perpetuate or normalize ableist attitudes, or to scrutinize or attack any individual annotators or original posters. This work is not intended as an ethical judgment or targeting of individuals, but rather an effort to improve AI alignment with the perspectives of autistic people.

\textbf{Citations and Acknowledgements:} When publishing findings, users should cite this dataset, acknowledging the work of its authors and the communities that provided the materials or data.

\textbf{Compliance with Regulations:} Researchers must comply with all relevant local, national, and international regulations and guidelines relating to data privacy and human subjects research where applicable.

\subsubsection{How to Request Approval}

\textbf{Commercial or Moderation Use:} If you intend to use the dataset for commercial purposes or automated content moderation, please submit a formal request, detailing:
\begin{itemize}
    \item Project objectives
    \item Potential for data use and distribution
    \item Mechanisms to ensure ethical application and protection of the data
    \item The impact of the project, and its target end-users
\end{itemize}

\textbf{Contact the First Author:} All requests and inquiries should be directed to the first author, as listed in the dataset documentation or project website, available here \footnote{https://nrizvi.github.io/AUTALIC.html}.

\subsubsection{Liability and Disclaimer}
The dataset is provided “as is,” without any guarantees regarding completeness, accuracy, or fitness for a particular purpose, especially outside of the U.S. context, for multi-media posts, or discussions on other platforms outside of Reddit.

\textbf{User Responsibility:} Users bear the responsibility for ensuring their use complies with these guidelines, as well as any applicable laws and ethical standards.

By accessing and using this dataset, you acknowledge that you have read and agreed to these Guidelines for Responsible Use, and that you understand the conditions under which the dataset may be utilized for your research or projects.

\subsection{Annotator Orientation}
\label{AO}
\subsubsection{Introduction}
This subsection provides an overview of the annotation orientation session conducted for \textsc{Autalic}. The goal is to ensure annotators understand the history and contemporary examples of anti-autistic ableism, the importance of neurodiversity, and the different ways in which anti-autistic speech may manifest in text. Given the sensitive nature of this work, annotators are advised that they may encounter discussions involving ableist language, violence, self-harm, and suicide mentions.

\subsubsection{Understanding Anti-Autistic Ableism}
Anti-autistic ableism is the discrimination and devaluation of autistic individuals based on neuronormative standards. A striking example includes cases where caretakers harm autistic individuals due to societal stigma ~\cite{disabilitymemorialDisabilityMourning}. The historical roots of such bias date back to Nazi-era eugenics research, where Hans Asperger categorized autistic individuals as either “useful” or “unfit,” reinforcing a harmful hierarchical perception of autism \cite{thetransmitterEvidenceTies}.

\paragraph{Neuronormativity}
Neuronormativity is the societal belief that neurotypical cognition is the default and that neurodivergence is an abnormality ~\cite{huijg2020neuronormativity}. This belief system marginalizes autistic individuals and contributes to discrimination in various aspects of life, including education, employment, and social interactions.

\paragraph{Deficit-Based Approaches and Their Harms}
Traditional medical models frame autism as a disorder requiring intervention or treatment ~\cite{kapp2013deficit, kapp2019social}. This perspective has led to:
\begin{itemize}
    \item Increased exposure to violence and self-harm risk
    \item Social exclusion and stigmatization
    \item Internalized ableism and lower self-esteem
\end{itemize}

\paragraph{Benevolent Ableism}
Benevolent ableism refers to actions or attitudes that, while seemingly supportive, reinforce autistic individuals as “less than” neurotypicals ~\cite{nario2019hostile}. Examples include organizations like \textit{Autism Speaks}, which promote awareness campaigns that fail to center autistic voices ~\cite{rosenblatt2022autism}. The use of symbols such as the puzzle piece is an example of this issue, as it implies that autism is a mystery to be solved rather than a valid identity.

\subsubsection{The Neurodiversity Movement}
The neurodiversity paradigm challenges the medical model by recognizing neurological variations as a natural and valid part of human diversity ~\cite{walker2014neurodiversity}. Symbols such as the rainbow infinity sign inspired by the LGBTQ Pride flag have emerged from within the community to counter external narratives that frame autism as a deficit ~\cite{kattari2023infinity}.

\paragraph{Community Perspectives}
Autistic individuals often reclaim language and challenge neuronormative narratives. Important considerations for annotation include:
\begin{itemize}
    \item Identity-first language (e.g., “autistic person” instead of “person with autism”) is preferred by the majority of autistic adults in the United States ~\cite{taboas2023preferences}
    \item Community-adopted terminology such as \textit{Aspie} (a self-identifier used by some autistic individuals)
\end{itemize}

\subsubsection{Annotation Tasks and Procedures}
In this section, we provide an overview of the annotation task along with video examples of the process.

\paragraph{Common Annotation Challenges}
Annotators should exercise careful judgment when evaluating phrases. For example:
\begin{itemize}
    \item Statements such as \textit{“That’s so autistic”} require contextual interpretation.
    \item The phrase \textit{“This vaccine causes autism”} is categorized as ableist due to its history in promoting autism stigma.
    \item The subtle difference between \textit{“I am not autistic”} and \textit{“At least I am not autistic”} changes the meaning and must be carefully assessed.
\end{itemize}

\subsubsection{Ethical Considerations and AI Bias}
\paragraph{Challenges in Hate Speech Detection}
Research indicates that many existing AI models misclassify disability-related discourse as toxic, even when the content is neutral or positive ~\cite{narayanan-venkit-etal-2023-automated, venkit-etal-2022-study, gadiraju_offensive_23, gamaartificially}. Specific issues include:
\begin{itemize}
    \item AI models exhibit over-sensitivity to disability-related discussions, frequently labeling them as harmful.
    \item AI models are more confident in detecting ableism when using \textit{person-first language} (e.g., “ableist toward autistic people”) than \textit{identity-first language} (e.g., “anti-autistic”). *
\end{itemize}
*\textit{these are results from our preliminary study}

\paragraph{Project Overview}
This project seeks to mitigate biases in AI hate speech detection by:
\begin{itemize}
    \item Training models using annotations informed by the neurodivergent community.
    \item Ensuring that AI does not misclassify community discourse as hate speech.
    \item Recognizing the distinction between hate speech and reclaimed terminology within the autistic community.
\end{itemize}

\subsubsection{Resources}
In this section, we provide resources such as our guidelines that contain a glossary to refer to or modify as needed. The terms in the glossary are those commonly used in neurodiversity discourse online.

\subsubsection{Conclusion}
The annotation orientation session is designed to equip annotators with the necessary knowledge to responsibly and accurately classify anti-autistic hate speech. By following the annotation guidelines and considering the broader socio-historical context, annotators contribute to the development of AI models that better serve neurodivergent individuals. 

\subsection{Search Keywords}
This list of terms in Table~\ref{tab:searchterms} were used to identify target sentences on Reddit. The number of target sentences containing each term is included.

\begin{table}
    \centering
    \begin{tabular}{cc} 
    \toprule
\bfseries Word & \bfseries Sentence Count \\ \midrule
autis*	& 1,221 \\ 
ASD	& 226 \\ 
disabilit* & 184 \\ 
aspergers & 173 \\ 
ABA	& 167\\ 
neurotyp*	& 158 \\ 
aspie* & 144 \\ 
neurodiver*	& 103 \\ 
AuDHD & 99 \\ 
disable*	& 93 \\ 
autism speaks	& 66 \\ 
stupid*	& 56 \\ 
a11y	& 34 \\ 
NT 	& 27 \\ 
retard*	& 25 \\ 
idiot*	& 18 \\ 
actually autistic	& 13 \\ 
autism intensifies	& 6 \\ 
ND 	      & 5 \\ 
autie* & 2 \\ \bottomrule
    \end{tabular}
    \caption{The keywords included in our Reddit search and the number of sentences containing each term in the \textsc{Autalic} dataset.}
    \label{tab:searchterms}
\end{table}

\subsection{Glossary (Excerpt)}
Below is an example of terms in our glossary. To view the full glossary, contact the first-author of this paper. This resource was created in collaboration with our annotators to define unfamiliar terms such as slang, medical abbreviations, laws, and other specialized language that they may encounter during the annotation process. In total, it contains 34 words.

\begin{table}[h]
    \centering
    \begin{tabular}{p{2cm} p{5cm}}
        \hline
        \textbf{Term} & \textbf{Definition} \\
        \hline
        AuDHD & A combination of autism and ADHD ~\cite{owensadhd}. \\
        Au, Âû & Used by autistic individuals to self-identify as autistic ~\cite{AU}. \\
        ND & Neurodivergent, neurodiverse, or neurodiversity  ~\cite{Greally_2021}
    \end{tabular}
    \caption{Glossary of specialized terms used during annotation.}
    \label{tab:glossary}
\end{table}

\subsection{Challenging Cases}
 Here, we show examples of sentences with high disagreements among our annotators, along with their own notes on the aspects of each sentence that made it difficult to classify:
 
\textbf{Example 1}  
\begin{quote}
    “ABA can be very beneficial to low-functioning autists who can't speak, use the bathroom, perform basic tasks, etc.”
\end{quote}
Promotes ABA and the “low-functioning” designation.
\textbf{Example 2}  
\begin{quote}
    “Autism can also lead to social issues like the autistic person saying the wrong thing at the wrong time with absolutely no utility involved.”
\end{quote}
Judgment that the autistic way is socially wrong.

\textbf{Example 3}  
\begin{quote}
    “Autism or ADHD or otherwise, you still have to learn basic etiquette.”
\end{quote}
Specific to cultural context: Implies NT (neurotypical) etiquette is required, but no expectation for NT to learn ways of relating that are natural to autistic people.

\textbf{Example 4}  
\begin{quote}
    “Basically, right after doing so much research, integrating with the autistic community, and accepting ASD as a part of myself, I was back to square one—left feeling like an idiot and immensely confused.”
\end{quote}
Slur against cognitive/intellectual disability, negativity associated with autistic identity, and medicalization of identity.

\subsection{Self-Agreement Scores}
\begin{figure}
   \centering
   \includegraphics[width=0.95\linewidth]{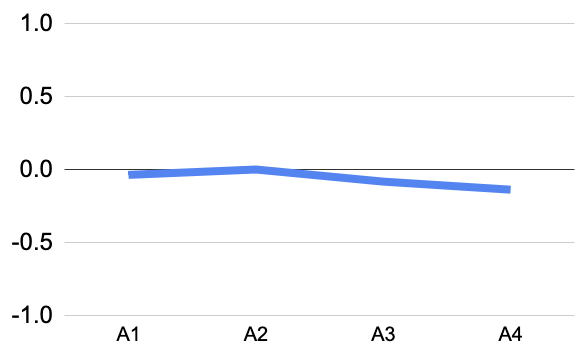}
   \caption{The self-agreement scores among annotators in a preliminary study highlight the difficulty of this task.}
   \label{fig:self-agreement}
\end{figure}

In preliminary studies, we examined different labeling schemes for this task to identify the most efficient and effective annotation strategy. Our experiments revealed high levels of self-disagreement among annotators, as shown in Figure ~\ref{fig:self-agreement}. The observed scores (M = -0.06, SD = 0.06) highlight the difficulty of the task and provide a meaningful baseline for comparison. Notably, our own annotation scores for \textsc{Autalic} were higher (M = 0.21, SD = 0.09), suggesting major improvement.

\subsection{Annotation Platform}
\begin{figure}[h]
    \centering
    \includegraphics[width=0.99\linewidth]{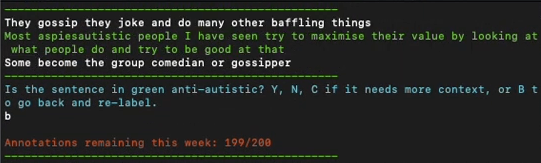}
    \caption{An example of an annotation task on our platform containing the target sentence (green) and contextual sentences (white).}
    \label{fig:platform}
\end{figure}

Figure ~\ref{fig:platform} shows an example of an annotation task on our platform with contextual sentences. 







\end{document}